\documentclass[conference]{IEEEtran}
\IEEEoverridecommandlockouts
\usepackage{cite}
\usepackage{amsmath,amssymb,amsfonts}
\usepackage{dsfont}
\usepackage{algorithmic}
\usepackage{graphicx}
\usepackage{textcomp}
\usepackage{xcolor}
\usepackage{hyperref}
\usepackage{multirow}
\def\BibTeX{{\rm B\kern-.05em{\sc i\kern-.025em b}\kern-.08em
    T\kern-.1667em\lower.7ex\hbox{E}\kern-.125emX}}
\begin{document}

\title{Uncertainty-aware data assimilation through variational inference\\
}

\author{\IEEEauthorblockN{Anthony Frion}
\IEEEauthorblockA{\textit{Institute of Coastal Systems} \\
\textit{Helmholtz-Zentrum Hereon}\\
Geesthacht, Germany \\
anthony.frion@hereon.de}
\and
\IEEEauthorblockN{David S Greenberg}
\IEEEauthorblockA{\textit{Institute of Coastal Systems} \\
\textit{Helmholtz-Zentrum Hereon}\\
Geesthacht, Germany \\
david.greenberg@hereon.de}
}

\maketitle

\begin{abstract}
Data assimilation, consisting in the combination of a dynamical model with a set of noisy and incomplete observations in order to infer the state of a system over time, involves uncertainty in most settings. Building upon an existing deterministic machine learning approach, we propose a variational inference-based extension in which the predicted state follows a multivariate Gaussian distribution. Using the chaotic Lorenz-96 dynamics as a testing ground, we show that our new model provides well calibrated predictions, and can be integrated in a wider variational data assimilation pipeline in order to achieve greater benefit from longer data assimilation windows. Our code is available at \url{https://github.com/anthony-frion/Stochastic_CODA}.
\end{abstract}


\section{Introduction}
\label{sec:intro}

In many geoscience problems, one has access to an accurate description of a dynamical system $\mathcal{M}$ of system states $\mathbf{x}_t \in \mathbb{R}^n$, but the capacity for directly measuring the full state is lacking. Data assimilation uses partial information from observations $\mathbf{y}_t$ to estimate $\mathbf{x}_t$ using the state-space equations:
\begin{align}
    \label{eq:model_equation}
    &\mathbf{x}_{t+1} = \mathcal{M}(\mathbf{x}_t) + \boldsymbol{\eta}_t, \\
    &\mathbf{y}_t = \mathcal{H}_t(\mathbf{x}_t) + \boldsymbol{\epsilon}_t.
\end{align}
The observation operator $\mathcal{H}_t$ relates $\mathbf{x}_t$ to $\mathbf{y}_t$, and in general cannot be inverted since $\mathbf{y}_t$ is typically much lower dimensional than $\mathbf{x}_t$. $\boldsymbol{\eta}_t$ and $\boldsymbol{\epsilon}_t$ respectively denote model and observation errors, often assumed to follow centered Gaussian distributions that do not depend on time $t$.

In this context, we seek to estimate the posterior probability distribution $p_\mathbf{x}(\mathbf{x}_{1:T} | \mathbf{y}_{1:T})$ of the state $\mathbf{x}_{1:T}$ given a set of corresponding observations $\mathbf{y}_{1:T}$. This data assimilation task is an inverse problem~\cite{aster2018parameter}, where the state equation~\eqref{eq:model_equation} is used (alongside a prior distribution of $\mathbf{x}_1$) to build a prior on $p_\mathbf{x}(\mathbf{x}_{1:T})$. Popular classes of methods for solving it include Kalman smoothers~\cite{gelb1974applied} and ensembles thereof~\cite{evensen2000ensemble}, as well as variational methods such as 4D-Var~\cite{le1986variational}. 
Many recent papers have proposed machine learning-based approaches (e.g.~\cite{fablet2021learning, farchi2021using, brajard2021combining, frerix2021variational, rozet2023score, wang2024four}), often with deterministic outputs (i.e. estimating only the maximum a posteriori of $p_\mathbf{x}(\mathbf{x}_{1:T} | \mathbf{y}_{1:T})$). 
While most of these methods rely on the supervised training of a neural network using ground truth state trajectories $\mathbf{x}_{1:T}$ or analysis performed by a classical assimilation algorithm, we leverage a recently proposed unsupervised learning method~\cite{zinchenko2024combined} which trains a model directly from noisy and incomplete observations $\mathbf{y}_{1:T}$ with no access to the corresponding states $\mathbf{x}_{1:T}$.

\section{Variational CODA}
\label{sec:var_CODA}

The neural network model $G_\theta$ from~\cite{zinchenko2024combined}, named Combined Optimization of Dynamics and Assimilation (CODA), takes as input a window of observations $\mathbf{y}_{t-w:t+w}$ 
and returns an estimate $\hat{\mathbf{x}}_t = G_\theta(\mathbf{y}_{t-w:t+w})$ 
of the most likely state at the center of the window given these observations. 
The parameters $\theta$ of the model are trained with the following unsupervised loss function:
\begin{multline}
\label{eq:CODA_loss}
    L(\theta) =
    \mathbb{E}_t \bigr[
    \sum_{i=0}^h ||
    \mathbf{y}_{t+i}
    - \mathcal{H}_{t+i} \circ \mathcal{M}^{(i)}(\hat{\mathbf{x}}_t)  ||^2 \\+
     \lambda ||\hat{\mathbf{x}}_{t+h} - \mathcal{M}^{(h)}(\hat{\mathbf{x}}_t)||^2\bigr],
\end{multline}
where $\circ$ denotes composition of functions, and $h$ is a hyperparameter representing the horizon of the prediction.
The first term is an observation error, measuring the agreement of the advancement of the predicted $\hat{\mathbf{x}}_t$ by $i$ time steps with the corresponding observations $\mathbf{y}_{t+i}$, for $i$ from $0$ to $h$. The second term is a regularization promoting the self-consistency of $G_\theta$ by comparing its time-propagated prediction to a prediction made by itself with a similar window of observations $h$ time steps later. The hyperparameter $\lambda$ is fixed and enables adjustment of the relative weight of these two loss terms.

Here, we modify $G_\theta$ so that, instead of returning a pointwise estimate $\hat{\mathbf{x}}_t$ of the state, it returns the parameters $\boldsymbol{\mu}_t \in \mathbb{R}^n, \boldsymbol{\sigma}_t \in \mathbb{R}^n$, so that the diagonal Gaussian distribution $q_t(\hat{\mathbf{x}}_t) =\mathcal{N}(\boldsymbol{\mu}_t, \boldsymbol{\Sigma}_t)$\footnote{From here on, $\boldsymbol{\Sigma}_t$ is a diagonal matrix with diagonal coefficients $\boldsymbol{\sigma}_t$.} forms our variational posterior estimate:
\begin{equation}
\label{eq:stochastic_CODA}
    G_\theta(\mathbf{y}_{t-w:t+w}) = (\boldsymbol{\mu}_t, \boldsymbol{\sigma}_t).
\end{equation}
It should be noted that the choice of a diagonal Gaussian distribution is rather restrictive, but that using a nondiagonal covariance matrix would pose important difficulties such as increased computational cost and potential spurious correlations. To train this new variational model, we must adapt the loss in equation~\eqref{eq:CODA_loss} so that it can be applied to a probability distribution over $\mathbf{x}$, rather than a point estimate. For the first loss term, corresponding to the observation error, we can simply average over samples from $q_t$, like in the expected likelihood term of the evidence lower bound~\cite{kingma2013auto}. The second term is less straightforward, since we aim to compare two distributions: samples from $q_t$ that have been advanced by $h$ simulation time steps (we denote the resulting distribution as $q_{t\rightarrow t+h}$), and the future variational posterior $q_{t+h}$. Ideally, we would like to minimize the Kullback-Leibler divergence:
\begin{equation} \label{eq:dkl_CODA}
    D_\text{KL}\left(q_{t\rightarrow t+h}\parallel 
    q_{t+h} \right)
    = \mathbb{E}_{\hat{\mathbf{x}} \sim q_{t\rightarrow t+h}} \log \frac{q_{t\rightarrow t+h}(\hat{\mathbf{x}})}{q_{t+h}(\hat{\mathbf{x}})}.
\end{equation}
Unfortunately, we cannot evaluate~\eqref{eq:dkl_CODA}, since we can sample from $q_{t\rightarrow t+h}$ but not evaluate its density. Instead, we add the entropy of $q_{t\rightarrow t+h}$ as an additional term, resulting in the loss
\begin{multline}
\label{eq:loss_CODA_gaussian}
    L(\theta) =
    \mathbb{E}_{t, \hat{\mathbf{x}}_t \sim q_t} \bigr[
    \sum_{i=0}^h || 
    \mathbf{y}_{t+i} 
    - \mathcal{H}_{t+i} \circ \mathcal{M}^{(i)}(\hat{\mathbf{x}}_t) ||^2 \\ -
     \lambda \; \log q_{t+h}(\mathcal{M}^{(h)}(\hat{\mathbf{x}}_{t})) \bigr].
\end{multline}
This choice is a natural extension of equation~\eqref{eq:CODA_loss} since the mean squared error can be understood as a negative log-likelihood with an identity covariance matrix. It also encourages lower-entropy posteriors. Note that, to evaluate the loss of equation~\eqref{eq:loss_CODA_gaussian}, we must make two passes through our network $G_\theta$, as for equation~\eqref{eq:CODA_loss}.  
Besides, the choice of $\lambda$ in equation~\eqref{eq:loss_CODA_gaussian} is critical for the calibration of the uncertainty of our probabilistic model, as removing the associated term would encourage the predicted variances to tend to $0$, actually leading to a deterministic model in practice.

\section{Training a stochastic CODA model}
\label{sec:training}

Following the experiments of~\cite{zinchenko2024combined}, we work on the Lorenz-96 dynamical system~\cite{lorenz1996predictability}, a popular benchmark in data assimilation, which models the evolution of a meteorological quantity on a latitude circle (i.e. the domain is periodic). It is composed of $n$ variables $x_1, ..., x_n$, each evolving as
\begin{equation}
    \frac{dx_i}{dt} = (x_{i+1} - x_{i-2}) x_{i-1} - x_i + F.
\end{equation}
We use $n=40$ variables and a forcing parameter $F=8$, resulting in chaotic dynamics with a doubling time of approximately $0.42$ time units (i.e. 2.1 days). The system is numerically integrated with a time step of $\delta t = 0.01$ (i.e. 1.2 hours) using the Runge-Kutta 4 integration scheme. The observation operator $\mathcal{H}$ randomly masks 75\% of the variables at each time step. The remaining variables are observed with a standard Gaussian observation error $\boldsymbol{\epsilon} \sim \mathcal{N}(\mathbf{0}_n, \mathbf{I}_n)$.

Our primary evaluation metric is the continuous ranked probability score (CRPS)~\cite{gneiting2007strictly}, averaged over time and over the $n$ dimensions of the system. The CRPS is a proper scoring rule enabling to compare the cumulative distribution function $F$ of a (one-dimensional) predicted probability distribution with a pointwise ground truth state $x_*$ as follows:
\begin{equation}
    \text{CRPS}(F,x_{*}) = \int_{-\infty}^\infty [F(x) - \mathds{1}_{x \geq x_{*}}]^2 \text{d}x.
    \end{equation}
The CRPS is a generalization of the mean absolute error to stochastic predictions. Indeed, for an ensemble $(x_1, ..., x_M)$ of $M$ equiprobable member predictions, we have:
\begin{equation}
\label{eq:crps_mae}
    \text{CRPS} = \frac{1}{M}\sum_{i=1}^M |x_{*}-x_i| - \frac{1}{2}\frac{1}{M^2}\sum_{i=1}^M\sum_{j=1}^M |x_i - x_j|,
\end{equation}
which trivially reduces to the mean absolute error when $M=1$, i.e. when performing a deterministic prediction. Equation~\eqref{eq:crps_mae} can be used in practice to estimate the CRPS of any probability distribution by sampling from it.

As secondary metrics, we use the spread and skill of the predictions. The spread of a predicted distribution is its standard deviation, and its skill is the root mean squared error of its mean with respect to the ground truth state. 
The spread and skill are expected to match on average when an accurate and well-calibrated posterior distribution has been computed, and one can test for this with the spread-skill plot. This consists in binning the spread values (obtained with varying inputs) in a histogram and computing the associated skills for each of these bins. Two synthetic metrics that one can derive from it are the spread-skill ratio (SSRAT) and spread-skill reliability (SSREL). The SSRAT is defined as the global ratio between the spread and the skill, and it has an ideal value of $1$, with values above indicating underconfidence and values below indicating overconfidence. The SSREL is the sum of absolute differences between the binned spread and skill values. Thus, it depends on the binning process and has an ideal value of 0. The interested reader can refer to~\cite{haynes2023creating} for more extensive descriptions of the CRPS, SSRAT and SSREL.

We train stochastic models on three datasets, respectively built from trajectories integrated over $10^4$, $3 \times 10^5$ and $3 \times 10^6$ time steps. Therefore, we hereafter refer to them as the small, medium and big datasets.

The benchmarked models are:

\textbf{Variational.} The method from section~\ref{sec:var_CODA}.

\textbf{Dropout.} An adaptation of the deterministic CODA model from~\cite{zinchenko2024combined}, with the simple addition of dropout~\cite{srivastava2014dropout} on its output layer, randomly masking out neurons with a probability $p$ that is adjusted to minimize the CRPS and obtain calibrated uncertainties. Following~\cite{gal2016dropout}, we use dropout during both training and inference, which results in a stochastic behavior mimicking Bayesian neural networks~\cite{mackay1992practical}.

\textbf{Ensembling.} We independently train 5 dropout models, which differ only by their random initializations, as suggested by~\cite{lakshminarayanan2017simple}, and combine their predictions when testing.

\begin{table}[tb!]
\caption{Performance of three stochastic prediction methods, on three different dataset sizes. For CRPS and SSREL, lower values are better. For SSRAT, values closer to 1 are better.}
    \begin{center}
    {
    \begin{tabular}{|c|c|c|c|c|}
        \hline
        \multicolumn{2}{|c|}{Dataset size} & Small & Medium & Big \\
        \hline
        \multirow{3}{5em}
        {Variational}
        & CRPS & 0.295 & $\mathbf{0.195}$ & $\mathbf{0.168}$ \\
        & SSRAT & 0.731 & $\mathbf{1.035}$ & $\mathbf{1.000}$ \\
        &  SSREL & ${0.105}$ & $\mathbf{0.012}$ &  $\mathbf{0.010}$ \\
         
         \hline
        \multirow{3}{5em}
        {Dropout}
        & CRPS & ${0.283}$ & 0.206 & 0.187 \\
        & SSRAT & ${0.780}$ & 1.040 & 0.970 \\
        &  SSREL & ${0.103}$ & 0.074 &  0.067 \\
         \hline

        \multirow{3}{5em}
        {Ensembling}
        & CRPS & $\mathbf{0.246}$ & 0.197 & 0.176 \\
        & SSRAT & $\mathbf{1.075}$ & 1.075 & 1.093 \\
        &  SSREL & $\mathbf{0.064}$ & 0.073 &  0.062 \\
        \hline
    \end{tabular}
    }
    \end{center}
\label{tab:results}
\end{table}
\begin{figure}[tb]
\begin{minipage}[b]{1.0\linewidth}
  \centering
  \centerline{\includegraphics[width=8.5cm]{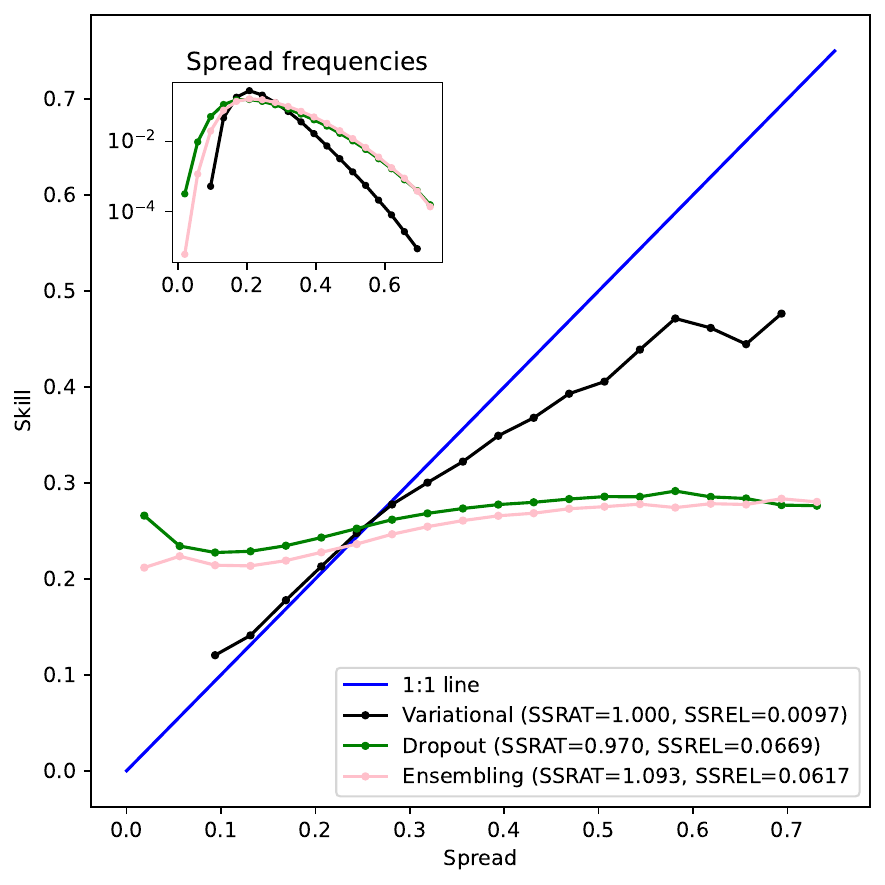}}
\end{minipage}
\caption{Spread-skill plots for the variational, dropout and ensembling models, trained on the big dataset. The inset spread frequencies plot indicates the relative weights of the dots of the main plot when computing the SSREL. The 1:1 line represents a perfect calibration.}
\label{fig:spread_skill}
\end{figure}

\begin{figure}[tb]
\begin{minipage}[b]{1.0\linewidth}
  \centering
  \centerline{\includegraphics[width=8.5cm]{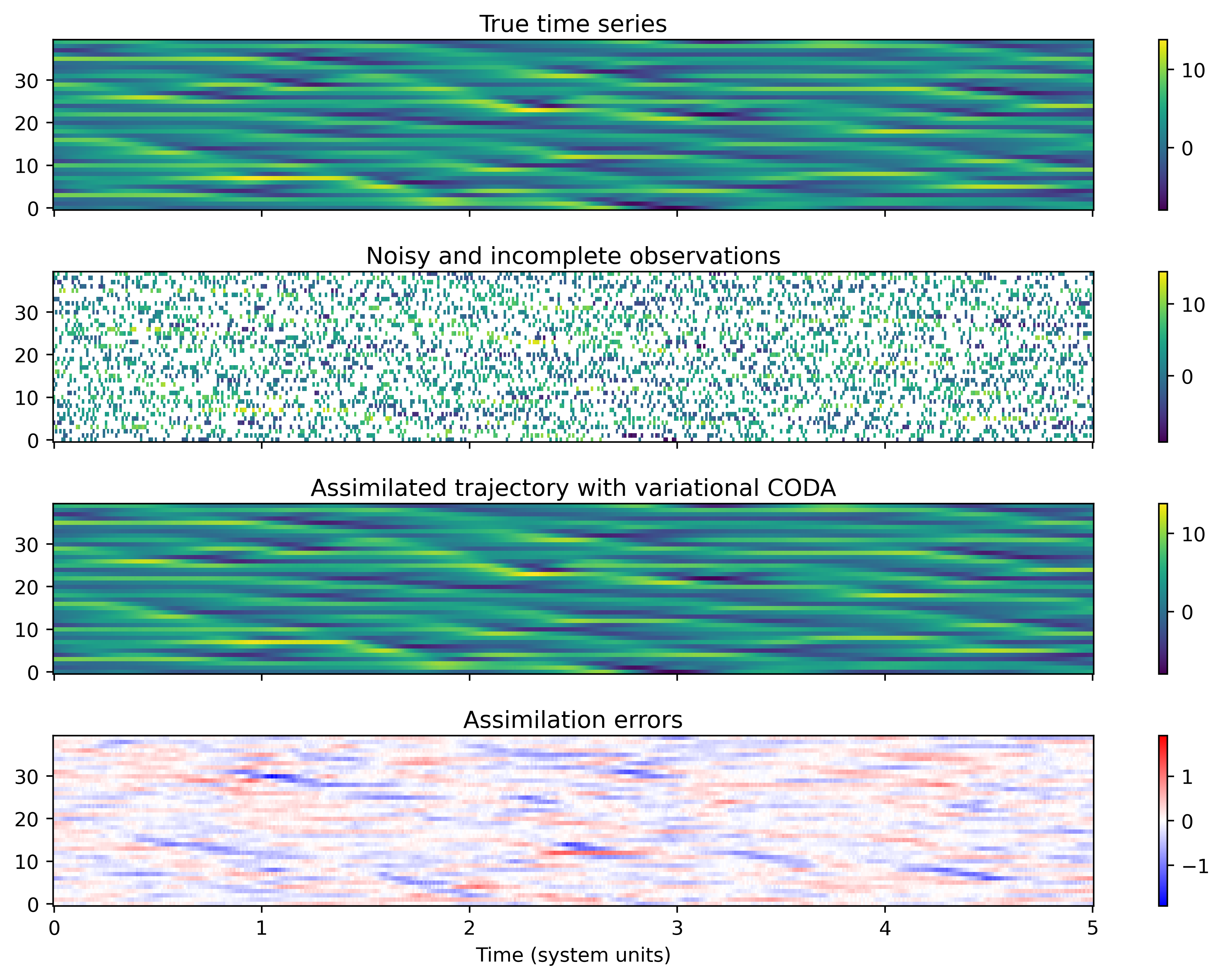}}
\end{minipage}
\caption{Visualization of a posterior mean estimated by our variational CODA method on the Lorenz-96 system. First row: ground truth state trajectory. Second row: Noisy and incomplete observations used by our method. Third row: resulting predicted trajectory with variational CODA. Fourth row: differences between the predicted and true states.}
\label{fig:stochastic_CODA_vis}
\end{figure}

The results of these 3 methods are summarized in table~\ref{tab:results}, and the spread-skill plots of their instances trained on the big dataset are shown on figure~\ref{fig:spread_skill}. In addition, in order to enable a qualitative assessment of the assimilation performance of our proposed variational method (trained on the big dataset), we display on figure~\ref{fig:stochastic_CODA_vis} an example of the estimated posterior mean of a short time series of Lorenz-96 states given a set of corresponding sparse and noisy observations.

From the results of table~\ref{tab:results} and figure~\ref{fig:spread_skill}, one can first see that the CRPS of all 3 benchmarked methods strongly decreases as the amount of training data increases. It further appears that ensembling 5 dropout models consistently reduces the CRPS with respect to a single dropout model. As can be seen from the fact that the spread-skill plot of the variational method is close to the 1:1 line in figure~\ref{fig:spread_skill}, this method obtains a significantly better SSREL than the other two methods when the training data is abundant. One can also note that ensembling dropout models improves the skill of the predictions but has a low impact on the spread, hence the increased SSRAT. 
While the SSREL of the variational method (and, to a lesser extent, of a single dropout model) clearly improves with larger amounts of training data, the same cannot be said for the ensemble of dropout models, for which there is no clear tendency. 
Overall, the variational method performs best except when trained on the smallest dataset. This can be explained by the ability of dropout layers to prevent overfitting in neural networks~\cite{srivastava2014dropout}, which is indeed most important when training on low amounts of data.
From these results, it seems likely that an ensemble of variational CODA models would perform even better than a single model, especially when re-tuning $\lambda$ in equation~\eqref{eq:loss_CODA_gaussian} to obtain a well calibrated ensemble.

\begin{table}[tb!]
\caption{Spread-skill ratio (SSRAT) of our model trained on the medium dataset, when using different values of the hyperparameter $\lambda$ in the loss function of equation~\eqref{eq:loss_CODA_gaussian}.}
    \begin{center}
    {
    \begin{tabular}{|c|c|c|c|c|}
        \hline
        Value of $\lambda$
        & 0 & 0.1 & 0.2 & 0.5 \\
        \hline
        SSRAT & 0.000 & 0.821 & 0.888 & 1.035 \\
        \hline
    \end{tabular}
    }
    \end{center}
\label{tab:lambda_sensitivity}
\end{table}
In order to gain further insight on the influence of the hyperparameter $\lambda$ in our loss function of equation~\eqref{eq:loss_CODA_gaussian}, we display in table~\ref{tab:lambda_sensitivity} the SSRAT values obtained when training stochastic CODA models on the medium dataset with different values of $\lambda$. As anticipated, the complete removal of the second term of the loss, corresponding to $\lambda = 0$, leads to a collapse of the predicted variance to $0$, i.e. a model with no spread, hence the SSRAT value of nearly $0$. Even for strictly positive values, it is clear that the spread is correlated with $\lambda$, which leads to increasing SSRAT values with increasing values of $\lambda$, although the sensitivity is relatively moderate.

\section{Using a trained model for 4D-Var}
\label{sec:4D-Var}

In the previous section, we have trained a stochastic CODA model that produces nearly perfectly calibrated diagonal Gaussian estimates of the state of a system, in the sense that their spread-skill ratio is nearly 1. This model, once trained, is much more computationally efficient than classical data assimilation methods. 
However, it computes an estimate of a state at time $t$ using a relatively small window of observations (e.g. 65 time steps for our model trained on the biggest dataset) and it does not directly simulate the known dynamics of the system at inference time. This means that one can expect improved performance with a costlier method, leveraging observations over much longer assimilation windows. Thus, in this section, we integrate a pre-trained instance of our stochastic CODA model into a classical weak-constraint 4D-Var computation scheme for assimilation on long observation windows. We show that this approach enables to improve the performance of deterministic data assimilation compared to a direct use of CODA, and more importantly compared to a similar 4D-Var scheme that does not leverage the CODA outputs.

Let us denote by $\mathbf{y}_{-w:T+w}$ a window of observations from the Lorenz-96 system, following the same characteristics as in section~\ref{sec:training}. 
We use a pre-trained variational CODA model to get initial predictions $(\boldsymbol{\mu}_t, \boldsymbol{\sigma}_t)$ on the corresponding state variables through equation~\eqref{eq:stochastic_CODA}, for every time t from $0$ to $T$.
Then, in its more general form, the assimilation cost that we aim to minimize in our 4D-Var scheme is expressed as
\begin{multline}
\label{eq:4DVar}
    J(\mathbf{x}_{0:T}) = \sum_{t=0}^T ||\mathcal{H}_t(\mathbf{x}_t) - \mathbf{y}_t||^2 + \alpha \sum_{t=1}^{T}||\mathbf{x}_t - \mathcal{M}(\mathbf{x}_{t-1})||^2 \\
    + \beta ||\mathbf{x}_0 - \boldsymbol{\mu}_0||_{\boldsymbol{\Sigma}_0}^2 + \gamma ||\mathbf{x}_T - \boldsymbol{\mu}_T||_{\boldsymbol{\Sigma}_T}^2,
\end{multline}
where $||\mathbf{x}||^2_\mathbf{A} = \mathbf{x}^\intercal\mathbf{A}^{-1}\mathbf{x}$ is a weighted Euclidean norm. $\alpha, \beta, \gamma$ are hyperparameters used to adjust the relative weights of the 4 terms of the cost. While $\alpha$ can take any positive value, we will only consider values of $0$ or $1$ for $\beta$ and $\gamma$, in order to either include or exclude the corresponding terms from the variational cost $J$. When $\beta = 1$ and $\gamma=0$, equation~\eqref{eq:4DVar} can be recognized as an instance of the classical weak-constraint 4D-Var cost (see e.g.~\cite{tr2006accounting}). 
It allows computing the maximum a posteriori of the distribution $p(\mathbf{x}_{0:T}|\mathbf{y}_{0:T})$ when the background prior distribution of the initial state is a Gaussian $\mathbf{x}_0 \sim \mathcal{N}(\boldsymbol{\mu}_0, \boldsymbol{\Sigma}_0)$, the model error in equation~\eqref{eq:model_equation} is $\boldsymbol{\eta}_t \sim \mathcal{N}(\mathbf{0}_n, \alpha^{-1} \mathbf{I}_n)$ and the observation error follows a standard Gaussian distribution. Setting $\beta=0$ to remove the prior term can be understood as using a uniform distribution on the domain of the state $\mathbf{x}$ as the prior distribution.
While the background prior on the initial state for each assimilation window is usually derived from analysis of a previous window, using stochastic CODA as a prior makes use of the first $w$ observations in the present window as well. Thus, these observations are used twice---once when setting up the prior, and again when minimizing the cost in equation~\eqref{eq:4DVar}.
Besides, when additionally setting $\gamma=1$, we make use of a ``foreground prior'', which we introduce analogously to the background prior in order to guide the assimilation cost at the end of the observation window. Note that using a foreground prior is not standard in variational data assimilation.

\begin{figure}[tb]

\begin{minipage}[b]{1.0\linewidth}
  \centering
  \centerline{\includegraphics[width=8.5cm]{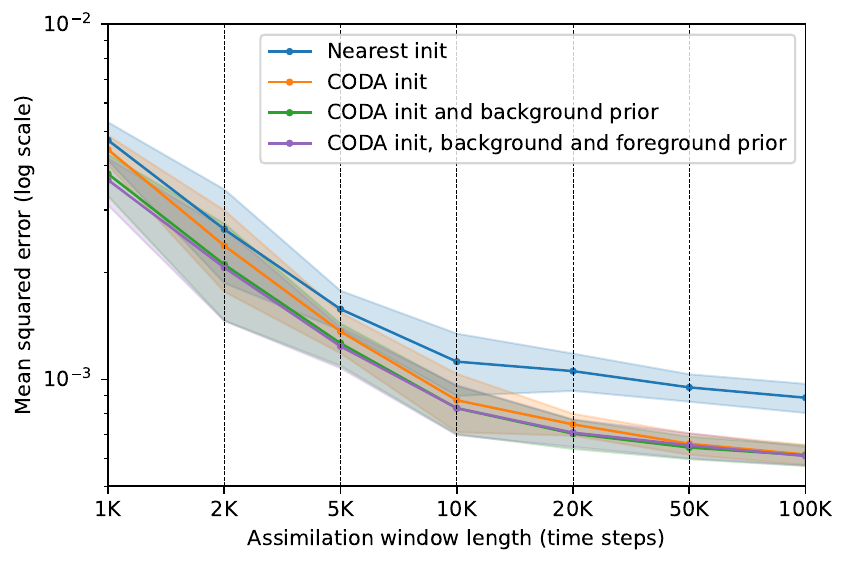}}
\end{minipage}
\caption{Mean squared error obtained by our 4 variants of 4D-Var, as a function of the assimilation window length. For each length, we use 10 different windows, and report the mean plus/minus standard deviation of the errors.}
\label{fig:4DVar_results}
\end{figure}

We test 4 variants of the assimilation procedure, on 7 different lengths $T$ of the observation window. For each value of $T$, we use 10 different sets of states and observations, and report the mean and standard deviation of the mean squared errors of each variant over these inverse problem instances.

The values of $T$ that we consider are [$1000$, $2000$, $5000$, $10^4$, $2\cdot 10^4$, $5 \cdot 10^4$, $10^5$]. We use $\alpha = 10^7$ in all experiments: a very high value since the dynamical model is in fact perfectly known here. The assimilation cost is minimized using automatic differentiation, and the optimizer is limited-memory BFGS, run for 5000 iterations with the default parameters from Pytorch. The 4 tested variants are:

\textbf{Nearest init.} A minimization of the cost from equation~\eqref{eq:4DVar} with $\beta = \gamma = 0$ and an initial value of $\mathbf{x}_{0:T}$ obtained by, for each variable and each time step, copying the available observation that is the closest in time for this variable. Like the other variants, this initialization makes use of a few observations outside of the assimilation window. This heuristic initialization was selected as a baseline as it surprisingly performed better than slightly more complex approaches such as linear interpolation and Cressman interpolation~\cite{cressman1959operational}.

\textbf{CODA init.} A variant that similarly sets $\beta = \gamma = 0$ but uses the mean predictions $\boldsymbol{\mu}_t$ of the pre-trained CODA model to initialize $\mathbf{x}_{0:T}$ in equation~\eqref{eq:4DVar}.

\textbf{CODA init and background prior.} Same as CODA init, but additionally using $\beta = 1$, so that the predicted mean and variance of the initial state are explicitly used to define the background prior term in the assimilation cost.

\textbf{CODA init, background and foreground prior.} Same as the previous variant, but additionally using $\gamma = 1$.

In figure~\ref{fig:4DVar_results}, we show the results obtained by these different variations. One can first observe that all of them strongly benefit from increasing assimilation window lengths. 
\textbf{CODA init} performs better than \textbf{Nearest init} in all cases, with a gap strongly increasing with the window length. This clearly shows that the weak-constraint 4D-Var method strongly benefits from being initialized with the fast and relatively accurate estimations that are provided by our stochastic CODA method. \textbf{CODA init and background prior} further enables significant improvements over \textbf{CODA init}, especially for the shorter window lengths, which shows that not only the mean of the distribution estimated by CODA is of interest, but also its associated well-calibrated uncertainties. Finally, the inclusion of a foreground prior with \textbf{CODA init, background and foreground prior} results in a marginal additional reduction of the mean squared error for shorter window sizes. Thus, both the background and foreground priors are most useful when the amount of observed data is relatively low.

\begin{figure}[tb]

\begin{minipage}[b]{1.0\linewidth}
  \centering
  \centerline{\includegraphics[width=8.5cm]{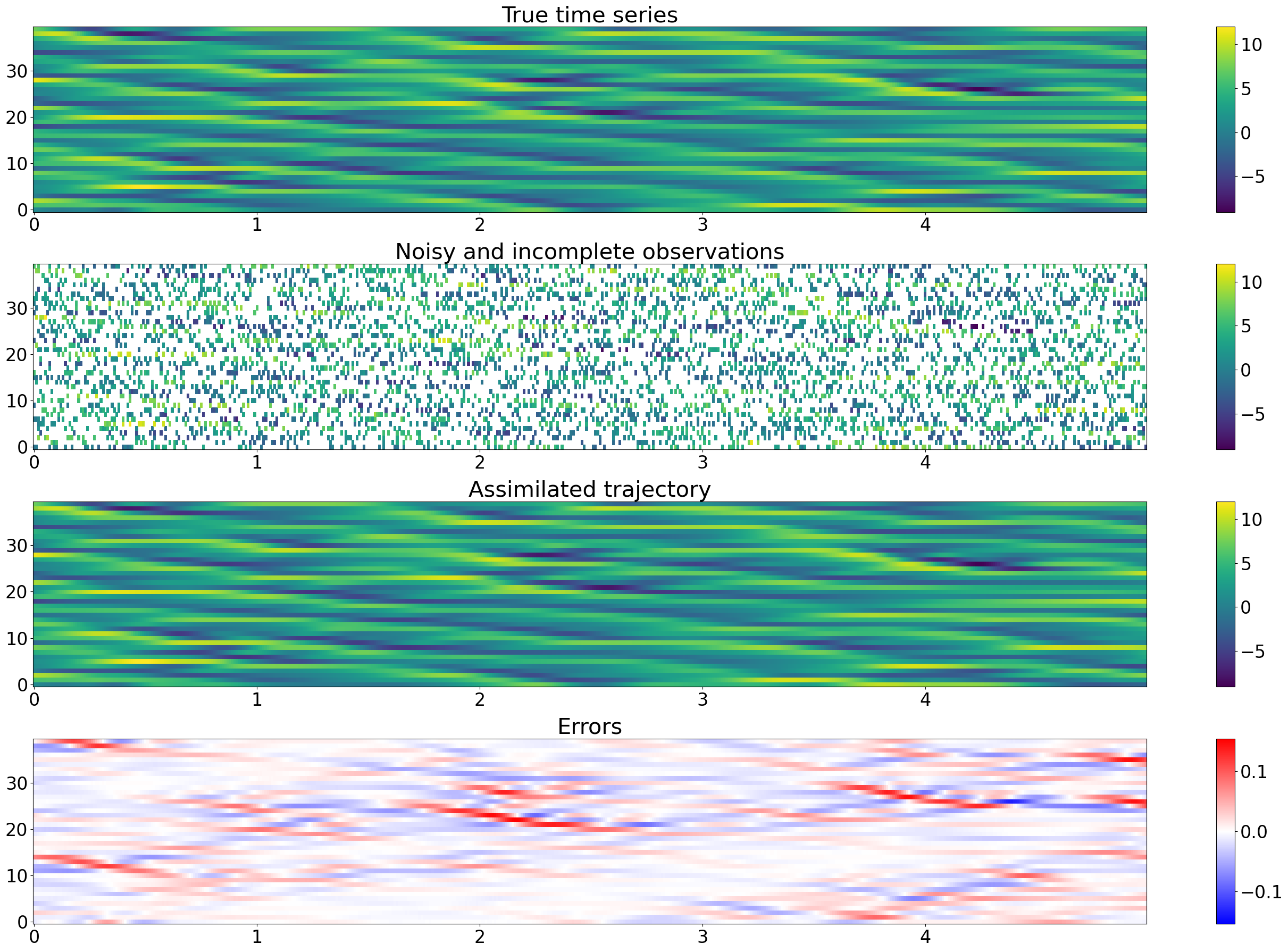}}
\end{minipage}
\caption{Visualization of a slice of an assimilation window with $T=10^5$ time steps. First row: ground truth state trajectory. Second row: Noisy and incomplete observations used in the 4D-Var cost. Third row: resulting predicted trajectory, using the cost of equation~\eqref{eq:4DVar} with $\beta=\gamma=1$. Fourth row: differences between the predicted and true states.
}
\label{fig:visu}
\end{figure}

In figure~\ref{fig:visu}, we show some results over a slice of one of our longest assimilation windows ($T=10^5$), assimilated with the CODA init, foreground and background prior approach. One can see that the ground truth is very well reconstructed using only sparse and noisy observations. In particular, comparing the scale of the errors to its equivalent in figure~\ref{fig:stochastic_CODA_vis} shows that minimizing the cost function of equation~\eqref{eq:4DVar} enables to significantly improve on the fast variational CODA estimate, which, as previously discussed, can be explained by the significantly higher computational cost and by the very long window of observations that this method has access to.

\section{Conclusion}
\label{sec:conclusion}

In this paper, we showed that it was possible to obtain nearly perfectly calibrated uncertainties in data assimilation with an unsupervised training of neural network for variational inference. We also demonstrated how such a pre-trained model can help improve the reconstruction ability of a classical weak-constraint 4D-Var method. However, in this experiment, we leverage stochastic estimates of the state in order to ultimately obtain deterministic predictions, and thus designing revised 4D-Var-like methods with stochastic outputs is a natural extension of this work. Furthermore, while we have only considered data assimilation with a perfectly known dynamical model, the CODA framework can also address more general tasks where the dynamics are partly unknown, and uncertainty-aware resolution of these tasks would certainly be of interest. Finally, we underline that our experiments are limited to the small-scale Lorenz-96 system, which is a popular testing ground for new data assimilation methods but remains largely simplified compared to operational settings, which are notably characterized by larger scales, heterogeneous real-world measurements with misspecified observation error distributions and complex background prior distributions. Future works should therefore study the adaptation of our methods for these more complex systems.


\bibliographystyle{IEEEbib}
\bibliography{refs.bib}

\end{document}